\documentclass[11pt,a4paper]{article}
\usepackage[hyperref]{naaclhlt2019}
\usepackage{times}
\usepackage{latexsym}

\usepackage{url}

\aclfinalcopy % Uncomment this line for the final submission
\def\aclpaperid{2085} %  Enter the acl Paper ID here
    
\title{CLEVR-Dialog:\\A Diagnostic Dataset for Multi-Round Reasoning in Visual Dialog}

\author{Satwik Kottur$^1$, 
Jos\'e M.F. Moura$^1$,
Devi Parikh$^{2,3}$, 
Dhruv Batra$^{2,3}$, 
Marcus Rohrbach$^2$\\ 
$^1$Carnegie Mellon University, $^2$Facebook AI Research, $^3$Georgia Institute of Technology\\
{\tt \{skottur,moura\}@andrew.cmu.edu,\{parikh,dbatra\}@gatech.edu,mrf@fb.com}}

\date{}

\usepackage{xspace}
\usepackage{url}
\usepackage{latexsym}
\usepackage{graphicx}
\usepackage{color}
\usepackage{xcolor}
\usepackage{xspace}
\usepackage{mysymbols}
\usepackage{enumitem}
\usepackage{microtype}
\usepackage{amsmath}
\usepackage{graphicx}
\usepackage{pslatex}
\usepackage{latexsym}
\usepackage{xspace}
\usepackage{multirow}
\usepackage{amsfonts}
\usepackage{multirow}
\usepackage{etoolbox}
\usepackage{subcaption}
\usepackage{microtype}
\usepackage{booktabs,arydshln}
\usepackage{floatrow}
\usepackage{sidecap}
\usepackage{flushend}
\newcommand{\ours}{CLEVR-Dialog\xspace}

\newcommand{\myquote}[1]{\emph{`#1'}}

\newcommand{\reffig}[1]{Fig.~\ref{#1}}
\newcommand{\refsec}[1]{Sec.~\ref{#1}}
\newcommand{\reftab}[1]{Tab.~\ref{#1}}
\makeatletter
\def\adl@drawiv#1#2#3{%
        \hskip.5\tabcolsep
        \xleaders#3{#2.5\@tempdimb #1{1}#2.5\@tempdimb}%
                #2\z@ plus1fil minus1fil\relax
        \hskip.5\tabcolsep}
\newcommand{\cdashlinelr}[1]{%
  \noalign{\vskip\aboverulesep
           \global\let\@dashdrawstore\adl@draw
           \global\let\adl@draw\adl@drawiv}
  \cdashline{#1}
  \noalign{\global\let\adl@draw\@dashdrawstore
           \vskip\belowrulesep}}
\makeatother

\usepackage{capt-of,calc}
\newsavebox{\figurebox}
\begin{document}
\maketitle
% main document
\begin{abstract}
%	\sk{
    Visual Dialog is a multimodal task of answering a sequence of questions 
    grounded in an image, using the conversation history as context.
    It entails challenges in vision, language, reasoning, and grounding. 
    However, studying these subtasks in isolation on large, real datasets 
    is infeasible as it requires prohibitively-expensive complete annotation of 
	the `state' of all images and dialogs.  
    
    We develop \ours, a large diagnostic dataset for studying multi-round reasoning in 
	visual dialog. 
    Specifically, we construct a \emph{dialog grammar} that is grounded in the 
    scene graphs of the images from the CLEVR dataset. 
	This combination results in a dataset where all aspects of the visual dialog 
	are fully annotated.     
    In total, \ours contains $5$ instances of $10$-round dialogs 
	for about $85k$ CLEVR images, totaling to $4.25M$ question-answer pairs.
    %In total, \ours contains $5$ instances of $10$-round dialogs 
	%for each of $70k$ (train) and $15k$ (val) CLEVR images, 
	%totaling to $3.5M$ (train) and $0.75M$ (val) question-answer pairs.
	
	We use \ours to benchmark performance of standard visual dialog models; 
    in particular, on \textit{visual coreference resolution} 
    (as a function of the coreference distance). 
	This is the first analysis of its kind for visual dialog models
	that was not possible without this dataset. 
    We hope the findings from \ours will help inform the development of future models for visual dialog.
    Our code and dataset are publicly available\footnote{\url{https://github.com/satwikkottur/clevr-dialog}}.
\end{abstract}

%% Various sections
\section{Introduction}
\begin{figure*}[t]
    \centering
    \includegraphics[width=\textwidth]{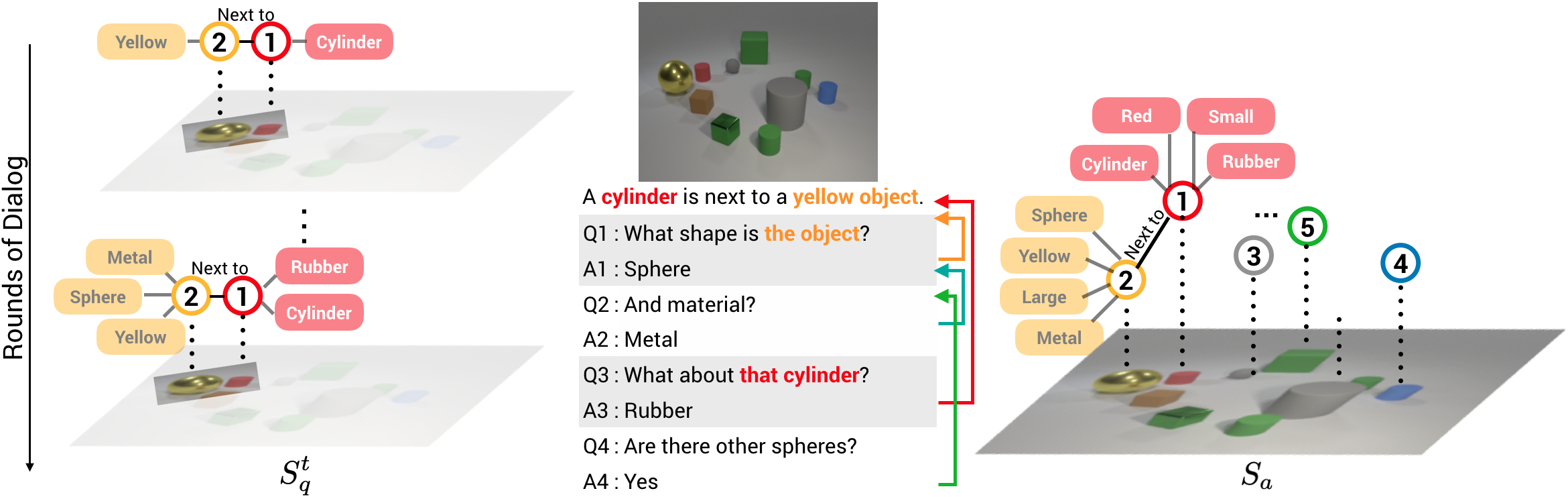}
    \caption{
    \ours: 
    we view dialog as 
communication between two agents -- an Answerer (A-er) who can `see'
the image $I$ and has the complete scene graph $S_a$ (far right),  
and a Questioner (Q-er), who does not `see' the image. 
A-er begins the dialog with a grounded caption 
(\myquote{A cylinder is next to a yellow object}). 
% This conveys some but not all information about $S_a$. 
The Q-er converts this caption into a partial scene graph 
$S^0_q$ (far left, top), follows up with a question grounded in $S^0_q$
(\myquote{What shape is the object?}), 
which the A-er answers, and the dialog progresses. 
Questions at round $t$ are generated based solely on 
    $S^t_q$, \ie, without looking at I or $S_a$, 
    which mimics real-life scenarios of visual dialog.
    }
    \label{fig:teaser}
\end{figure*}

The focus of this work is on intelligent systems that can 
\emph{see} (perceive their surroundings through vision), 
\emph{talk} (hold a visually grounded dialog), and 
\emph{reason} (store entities in memory as a dialog progresses, 
refer back to them as appropriate, count, compare, \etc). 
Recent works have begun studying such systems under the 
umbrella of \emph{Visual Dialog} \cite{visdial,guesswhat}, 
where an agent must 
answer a \emph{sequence} of questions grounded in an image. 
As seen in \reffig{fig:teaser}, this entails 
challenges in --
vision (\eg, identifying objects and their attributes in the image),
language/reasoning (\eg, keeping track of and referencing 
previous conversation via memory), 
and grounding (\eg, grounding textual entities in the image).

%%%%%%%%%%%%%%%%%%%%%%%%%%%%%%%%%%%%%%%%%%%%%%%%%%%%%%%%%%%%%%
% teaser figure
%%%%%%%%%%%%%%%%%%%%%%%%%%%%%%%%%%%%%%%%%%%%%%%%%%%%%%%%%%%%%%

In order to train and evaluate agents for Visual Dialog, 
\newcite{visdial} collected a large dataset 
of  human-human dialog on real images %from COCO~\cite{mscoco}, 
collected between pairs of workers on Amazon Mechanical Turk (AMT). 
While such large-scale realistic datasets enable 
new lines of research, it is difficult to 
study the different challenges (vision, language, reasoning, grounding) 
in isolation or to break down the performance of systems 
over different challenges to identify bottlenecks,  
because that would require prohibitively-expensive complete annotation of 
the `state' of all images and dialogs (all entities, coreferences, \etc).

In this work, we draw inspiration from \newcite{johnson2017clevr}, 
and develop a large diagnostic dataset---\ours---for 
studying and benchmarking multi-round reasoning in 
visually-grounded dialog. 
Each CLEVR image 
is synthetically rendered by a particular scene graph \cite{johnson2017clevr}  
and thus, is by construction exhaustively annotated. 
We construct a \emph{dialog grammar} that is grounded in 
these scene graphs. 
Specifically, similar to \newcite{das_iccv17}, we view dialog 
generation as 
communication between an Answerer (A-er) who can `see'
the image and has the complete scene graph (say $S_a$), 
and a Questioner (Q-er), who does not `see' the image and is trying to 
reconstruct the scene graph over rounds of dialog (say $S^t_q$).
As illustrated in \reffig{fig:teaser}, the dialog begins by A-er providing a 
grounded caption for the image, which conveys some but not all information 
about $S_a$. %, which forms $S_^{(0)}_p(I)$ for Q-er. 
The Q-er builds a partial scene graph $S^0_q$ based on the caption,
and follows up by asking questions grounded in $S^0_q$, which the A-er answers, and the dialog progresses.
Our dialog grammar defines rules and templates for constructing 
this grounded dialog.
Note that A-er with access to $S_a$ (perfect vision) exists
\textbf{only} during dialog generation to obtain ground truth answers.
While studying visual dialog on \ours, models are forced to answer
questions with just the image and dialog history 
(caption and previous question-answer pairs) as additional inputs.

In total, \ours contains $5$ instances of $10$-round dialogs 
for each of $70k$ (train) and $15k$ (val) CLEVR images, 
totaling to $3.5M$ (train) and $0.75M$ (val) question-answer pairs.
We benchmark several visual dialog models
on \ours as strong baselines for future work.

\begin{SCfigure}[][t]
    \centering
    \caption{Breakdown of performance by questions that depend on entire history
    (\textit{All}), require coreference resolution (\textit{Coref}),
    and are history-independent (\textit{None}).}
    \includegraphics[width=0.4\columnwidth]{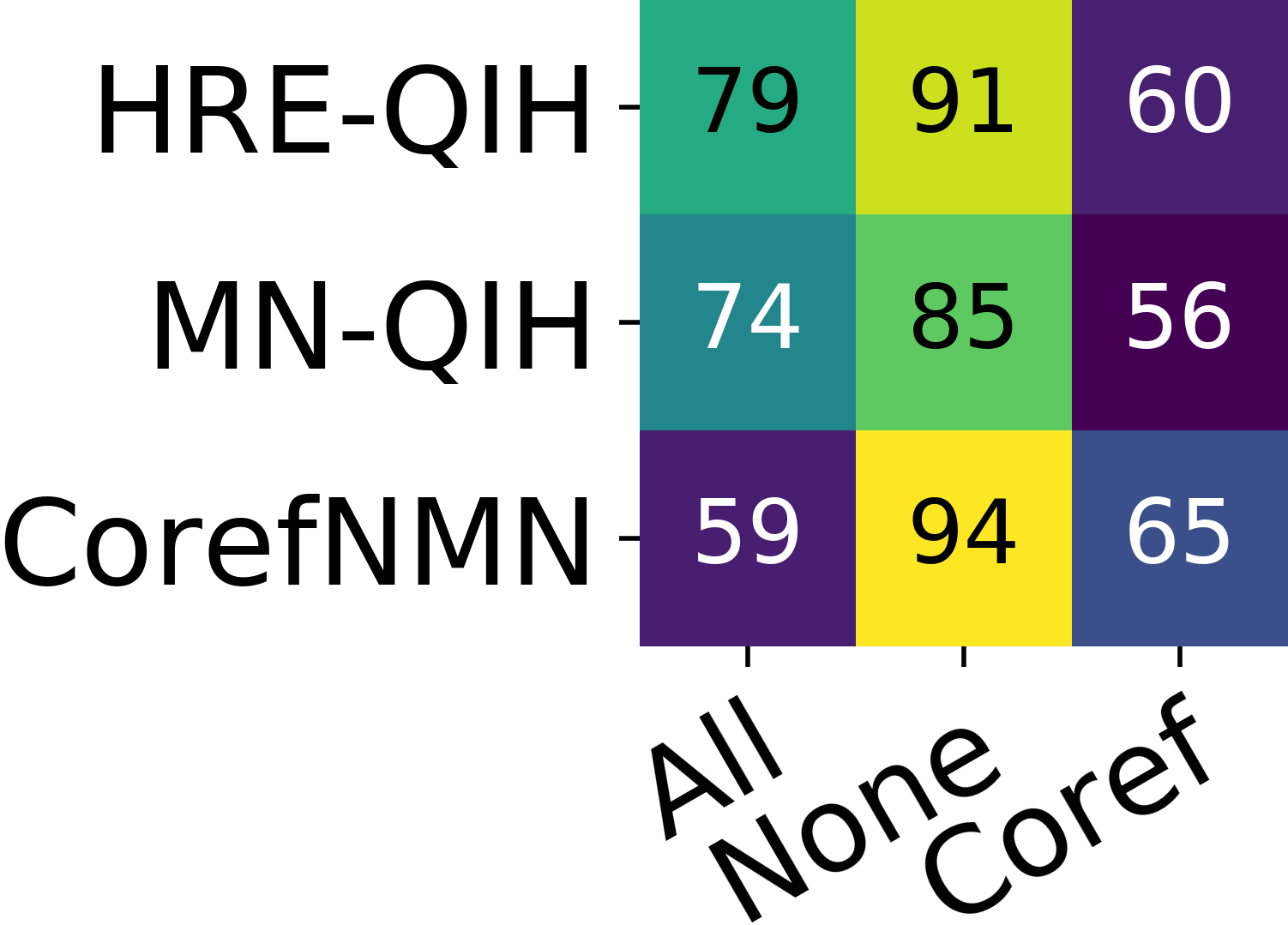}
    \label{fig:accuracy_coref_coarse}
\end{SCfigure}

The combination of CLEVR images (with full scene graph annotations) 
and our dialog grammar results in a dataset where all aspects of the visual dialog 
are fully annotated. We use this to study one particularly difficult 
challenge in multi-dialog visual reasoning -- 
of \textit{visual coreference resolution}.
A coreference arises when two or more phrases (\textit{coreferring phrases})
in the conversation refer to the same entity (\textit{referent}) in the 
image.
For instance, in the question \textit{`What about that cylinder?'} (Q3) 
from \reffig{fig:teaser},
the referent for the phrase \textit{`that cylinder'} can be inferred only 
after resolving the phrase correctly based on the dialog history, as there
are multiple cylinders in the image.
We use \ours to diagnose performance of different methods 
as a function of the history dependency (\eg., coreference distance---the number 
of rounds between successive mentions of the same object)
and find that the performance of a state-of-art model (CorefNMN) 
is at least 30 points 
inferior for questions involving coreference resolution compared to 
those which do not (\reffig{fig:accuracy_coref_coarse}), 
highlighting the challenging nature of our dataset.
This is the first analysis of its kind for visual dialog
that was simply not possible without this dataset.
We hope the findings from \ours will help inform 
the development of future models for visual dialog.

\section{Related Work}
% qualitative figure
\begin{figure*}[t]
	\centering
    \includegraphics[width=\textwidth]{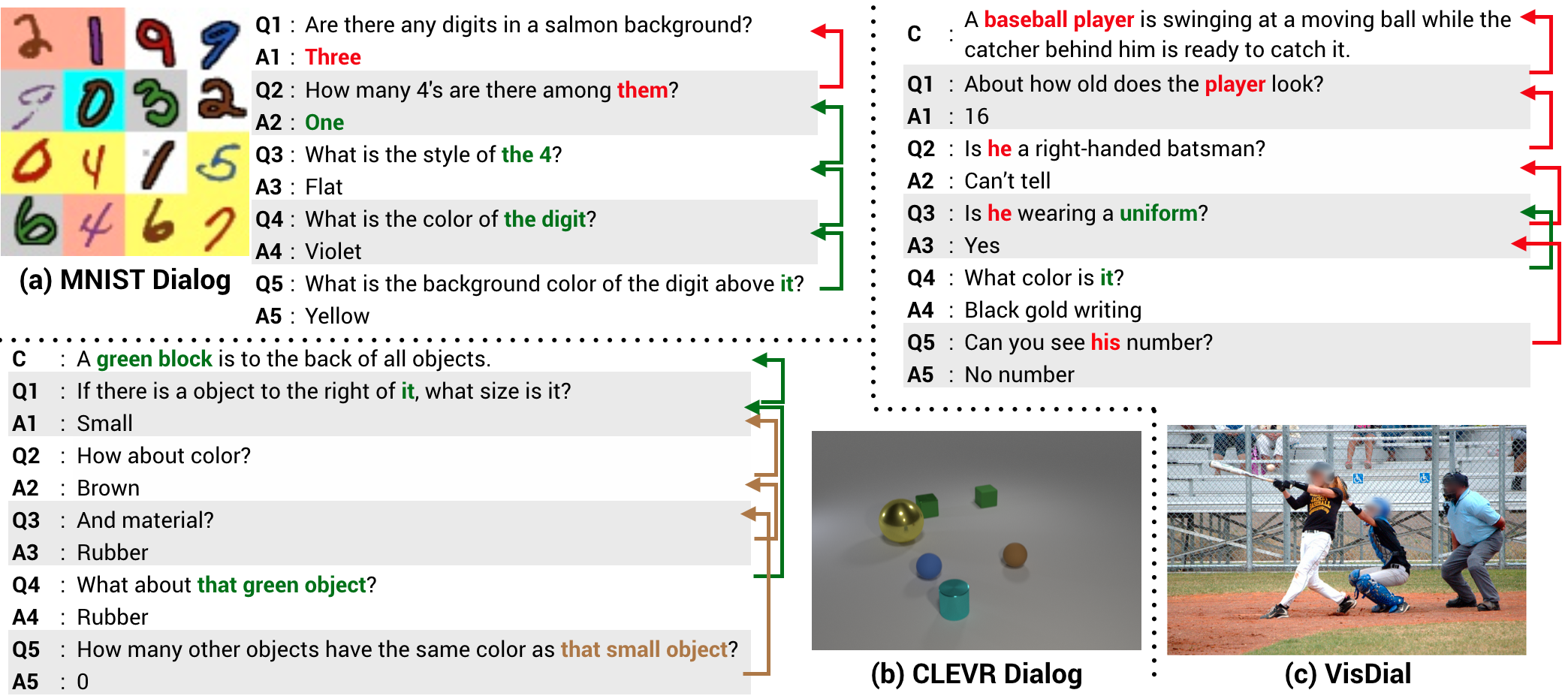}
    \caption{
    Example dialogs from MNIST Dialog, \ours, and VisDial, with coreference
    chains manually marked for VisDial and automatically extracted for 
    MNIST Dialog and \ours.
}
    \label{fig:qualitative}
\end{figure*}

\noindent
\textbf{Coreference Resolution}
	is a well studied problem in the NLP community \cite{Ng:2010:SNP:1858681.1858823,leeHLZ17,wisemanRS16,clark16emnlp,clark_manning_acl16}. 
   Our work focuses on \emph{visual} coreference resolution 
%   that adds an additional modality 
   -- the referent is now a visual entity 
   to be grounded in visual data.
   Several works have tackled visual coreference resolution in videos 
   \cite{ramanathan_eccv14,rohrbach17cvpr} and 3D data \cite{kong_cvpr14}, and have 
   introduced real image datasets for the same \cite{flickr30k}. 

\paragraph{Visual Dialog and Synthetic Datasets.}
We contrast \ours against four existing datasets:
%\noindent
(1) \textbf{CLEVR} \cite{johnson2017clevr} is 
    a diagnostic dataset for visual question answering (VQA) \cite{antol15iccv}
    on rendered images that contain objects like cylinders, cubes, \etc., 
    against a plain background (\reffig{fig:teaser}).
    While \ours uses the same set of images, the key difference 
    is that of focus and emphasis -- 
    the objective of CLEVR-VQA questions is to stress-test 
    spatial reasoning in 
    independent single-shot question answering; 
    the objective of \ours is to stress-test 
    temporal or multi-round reasoning over the dialog history.
(2) \textbf{CLEVR-Ref+}
    \cite{DBLP:journals/corr/abs-1901-00850} is a diagnostic dataset based on 
    CLEVR images for visual reasoning in referring expressions.
    \ours goes beyond CLEVR-Ref+, which focuses on grounding objects given a natural 
    language expression, and deals with additional visual and linguistic
    challenges that require multi-round reasoning in visual dialog.
(3) \textbf{MNIST-Dialog} \cite{paul2017visual} is a synthetic dialog dataset 
    on a grid of $4 \times 4$ stylized MNIST digits (\reffig{fig:qualitative}).
    While MNIST-Dialog is similar in spirit to \ours, 
    key difference is complexity -- 
    the distance between a coreferring phrase and its 
    antecedent is always 1 in MNIST-Dialog; in contrast, 
    \ours has a distribution ranging from $1$ to $10$. 
    (4) \textbf{VisDial} \cite{visdial} is a large scale visual dialog dataset
    collected by pairing two human annotators (a Q-er and an A-er)
    on AMT, built on COCO \cite{mscoco} images.
    VisDial being a large open-ended real dataset 
    encompasses all the challenges of visual dialog, 
    making it difficult to study and benchmark progress on 
    individual challenges in isolation.
\reffig{fig:qualitative} qualitatively compares MNIST-Dialog, \ours, and 
VisDial, and shows coreference chains (manually annotated for this VisDial 
example by us, and automatically computed for MNIST-Dialog and \ours). 
We can see that the coreference links in MNIST-Dialog 
are the simplest (distance always 1).
While coreferences in VisDial can be on a similar level of difficulty than \ours, the difficult cases are rarer in VisDial.

\section{\ours Dataset}
\label{sec:approach}
In this section, we describe the existing annotation for CLEVR images,
then detail the generation process for \ours, 
and present the dataset statistics in comparison to existing datasets.

\subsection{CLEVR Images}
Every CLEVR image $I$ has a full scene graph annotation,
$S_a$.
This contains information about all the objects in the scene, including
four major attributes $\{$\textit{color, shape, material, size}$\}$,
2D image and 3D world positions, and relationships 
$\{$\textit{front, back, right, left}$\}$ between these objects.
% \marcus{[Marcus: TODO: note at this place that we always refer to the closest object? see slack discussion]}
The values for the attributes are:
(a) \textit{Shape}---cylinder, cube, sphere;
(b) \textit{Color}---blue, brown, cyan, gray, green, purple, red, yellow;
(c) \textit{Size}---large and small; and finally
(d) \textit{Material}---metal and rubber.
We only use objects, attributes, and relationships.
% \marcus{it might clearer to say what we are not using: "2D image and 3D world positions,"!?}

\begin{figure*}[t]
	\centering
    \includegraphics[width=0.95\textwidth]{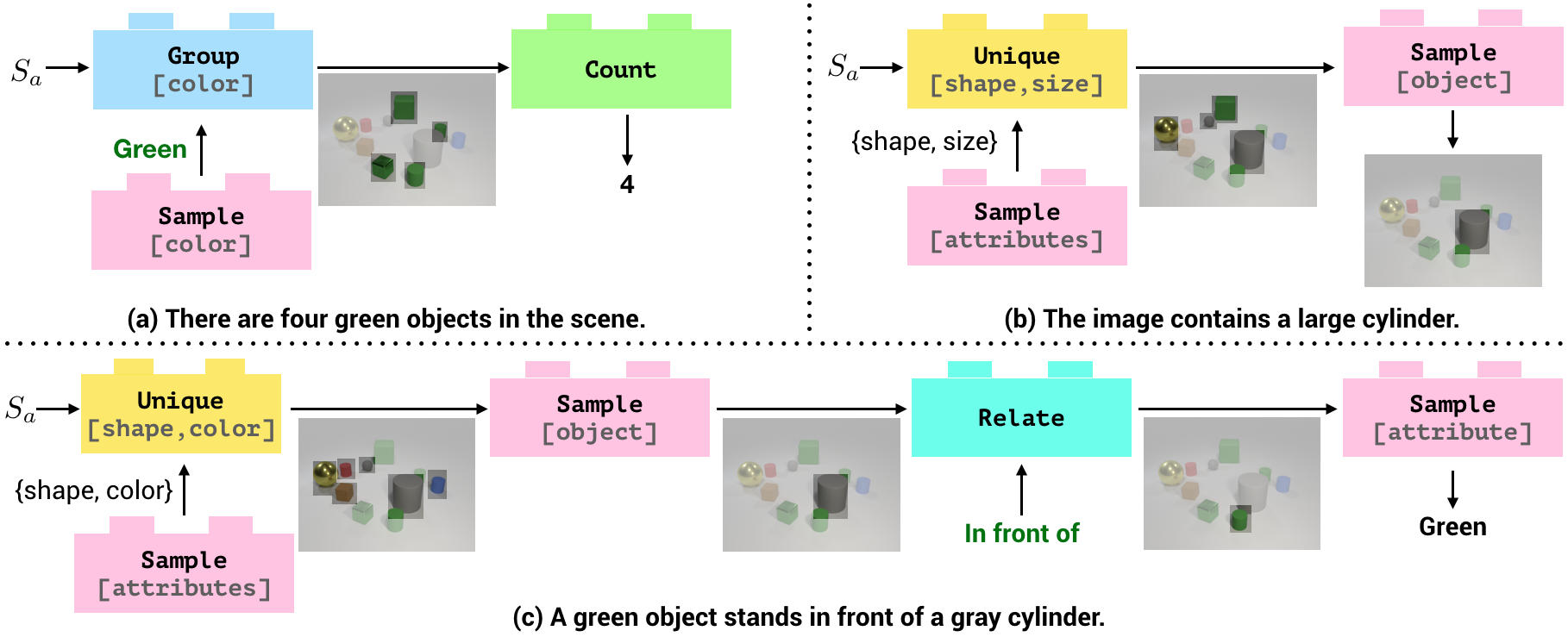}
    \caption{Usage of dialog grammar in caption generation.}
    %\marcus{Both in b, and c, "Sample[attributes]" results in "{shape,color}". It 
    %would be better to show possible variations.}\marcus{Is there no sample in (a) 
    %for "Green"?}}
    \label{fig:caption_generation}
\end{figure*}

\subsection{Dataset Generation}
An important characteristic of visual dialog that makes it 
suitable for practical applications is that the questioner 
does not `see' the image (because if it did, it would not 		
\emph{need} to ask questions). 
To mimic this setup, 
we condition our question generation at round $t$ only on 
the partial scene graph $S^t_q$ that accumulates 
information received so far from the dialog history (and not on $S_a$). 
Specifically, we use a set of caption $\{T^C_i\}$ and question
$\{T^Q_i\}$ templates (enumerated in \reftab{tab:templates}),
which serve as the basis for our dialog generation.
Each of these templates in turn consists of primitives,
composed together according to a generation grammar.
The nature and difficulty of the dataset is highly dependent on these 
templates, thus making their selection crucial.
In what follows, we will first describe these primitives, discuss how they are
used to generate a caption or a question at each round, and tie everything together
to explain dialog generation in \ours.
%%%%%%%%%%%%%%%%%%%%%%%%%%%%%%%%%%%%%%%%%%%%%%%%%%%%%%%%%%%%%%%%%%%%%%%%%%

\paragraph{Grammar Primitives.}
% Low priority: Supplement, okay for now.
%\marcus{[I find this subsection difficult to understand without examples]}
The templates used to generate captions and questions are composed of 
intuitive and atomic operations called primitives.
Each of these primitives can have different instantiations depending on a parameter,
and also take input arguments.
For example, \texttt{Filter} primitives filter out objects from an input set of
objects according to certain constraints.
In particular, \texttt{Filter[color](blue)} filters out blue objects from a given set of
objects, while \texttt{Filter[shape](sphere)} filters out all spheres.
In our work, we use the following primitives:
\begin{itemize}[leftmargin=0.15in,itemsep=-5pt]
\item \texttt{Sample}: sample an object/attribute, 
\item \texttt{Unique}: identify unique objects/attributes,
\item \texttt{Count}: count the number of input objects,
\item \texttt{Group}: group objects based on attribute(s),
\item \texttt{Filter}: filter inputs according to a constraint,
\item \texttt{Exist}: check for existence of objects,
\item \texttt{Relate}: apply a relation (\eg, \textit{right of}).
\end{itemize}
%%%%%%%%%%%%%%%%%%%%%%%%%%%%%%%%%%%%%%%%%%%%%%%%%%%%%%%%%%%%%%%%%%%%%%%%%%%
Note that each of these primitives inherently denotes a set of constraints,
which when failed leads to a reset of the generation process for the
current caption/question in the dialog.
For example, if the output of \texttt{Filter[color](blue)} is empty due to the
absence of blue objects in the input, we abort generation for the
current template and move on to the next template.

%%%%%%%%%%%%%%%%%%%%%%%%%%%%%%%%%%%%%%%%%%%%%%%%%%%%%%%%%%%%%%%%%%%%%%%%%%%
\paragraph{Caption Generation.}
The role of the caption is to \textit{seed} the dialog and initialize $S^0_q$.
   In other words, caption gives Q-er partial information about the image
   so that asking follow-up questions is possible.
   Because A-er generates the caption, it uses the full scene graph $S_a$.
   \reffig{fig:caption_generation} shows the caption grammar in action, producing
   three different captions for a given image.
   Consider the grammar for \reffig{fig:caption_generation}(c).
   First, \texttt{Sample[attributes]} produces \textit{\{shape, color\}} used by
   \texttt{Unique} to  select objects from $S_a$ with unique shape and color attributes.
   An object (gray cylinder) is then sampled from these using \texttt{Sample[object]}.
   Next, a relation (\textit{in front of}) is enforced via a \texttt{Relate} primitive
   leading to the green cylinder in front of the gray cylinder.
   Finally, \texttt{Sample[attribute]} samples one of the attributes to give us the
   caption, \textit{`A green object stands in front of a gray cylinder.'}
   
% As the dialog between Q-er and A-er is initiated by the caption, care must be
%taken to ensure it is \textit{informative enough} to spawn clarifying questions from
%the Q-er.
We carefully design four different categories of caption templates: 
(a) \texttt{Obj-unique} mentions an object with unique set of attributes in the image,
(b) \texttt{Obj-count} specifies the presence of a group of objects with common 
attributes,
(c) \texttt{Obj-extreme} describes an object at one of the positional extremes of 
the image (right, left, fore, rear, center),
(d) \texttt{Obj-relation} talks about the relationship between two objects along with
their attributes in a way that allows them to be uniquely identified in the complete
scene graph $S_a$.
In our work, the relationships are used in an immediate or closest sense,
\ie., a relation \textit{to the right of} actually means \textit{to the immediate right of}.
% \marcus{[TODO: note that the relationship is unique]}
\reftab{tab:templates} shows example captions.
% In contrast, MNIST Dialog does not have captions.}
%%%%%%%%%%%%%%%%%%%%%%%%%%%%%%%%%%%%%%%%%%%%%%%%%%%%%%%%%%%%%%%%%%%%%%%%%%%

\begin{figure*}[t]
	\centering
    \includegraphics[width=0.95\textwidth]{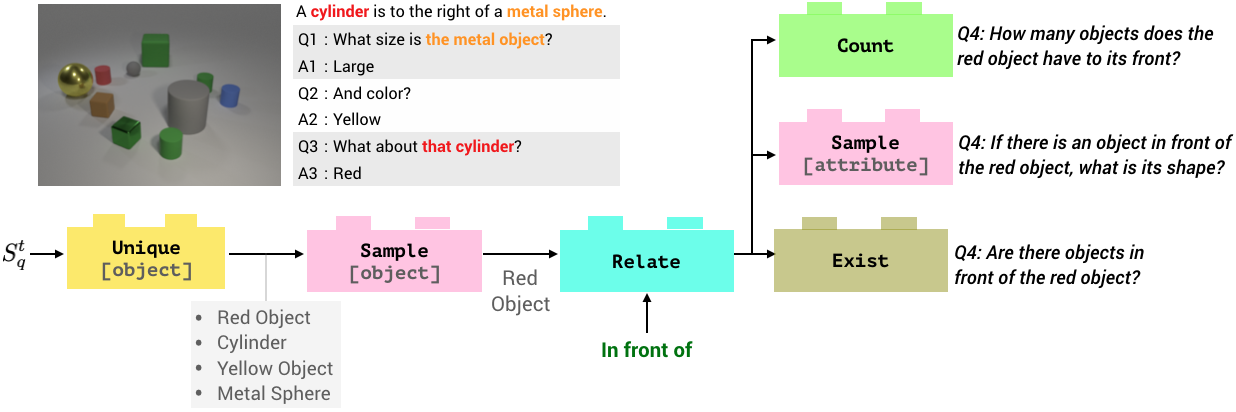}
    \caption{Usage of dialog grammar in question generation.}
    %\marcus{FIX image/dialog as discussed in slack.}}
    \label{fig:question_generation}
\end{figure*}

\begin{figure}[t]
	\centering
    \includegraphics[width=0.98\columnwidth]{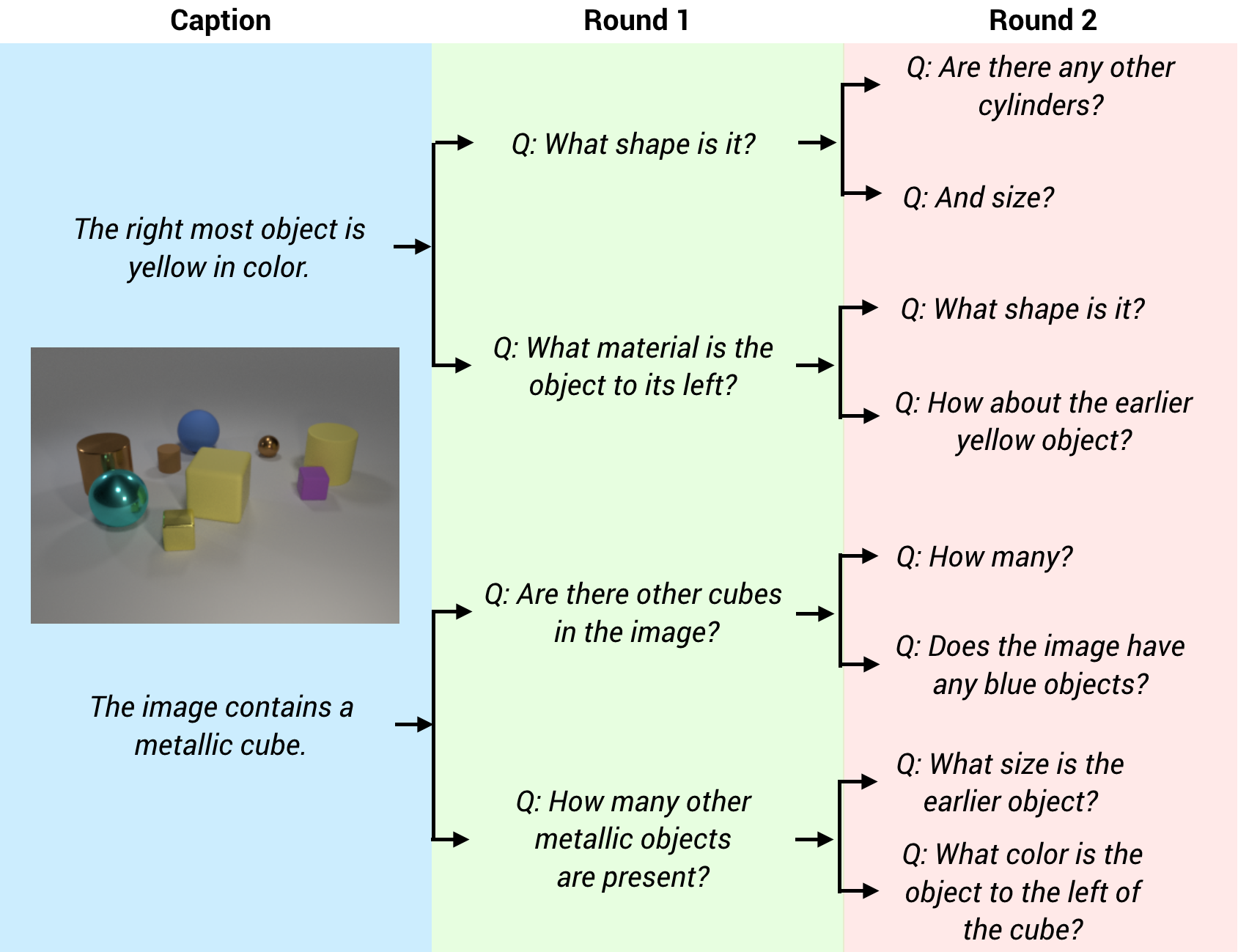}
    \caption{Dialog generation in \ours.
    At each round, all valid question templates are used to generate candidates for 
    the next question.
    However, only a few \textit{interesting} candidates (beams) are retained for 
    further generation, thus avoiding an exploding number of possibilities as 
    rounds of dialog progress.}
    \label{fig:dialog_generation}
\end{figure}
%%%%%%%%%%%%%%%%%%%%%%%%%%%%%%%%%%%%%%%%%%%%%%%%%%%%%%%%%%%%%%%%%%%%%%%%%%%
\paragraph{Question Generation.}
Unlike the caption, the questions are generated by the Q-er, having access only to a
partial scene graph $S^t_q$ at round $t$.
This $S^t_q$ is an assimilation of information from the previous rounds of the dialog.
The primitives in the question template therefore take $S^t_q$ as the input scene graph,
and the generation proceeds in a manner similar to that of the caption explained above.
As the dialog is driven by Q-er based on partial scene information, only a few 
questions are non-redundant (or even plausible) at a given round of the dialog.
To this end, the inherent constraints associated with the primitives now play a bigger
role in the template selection.

In this work, we experiment with three different categories of
question templates: 
(a) \textbf{Count} questions ask for a count of objects in the image 
satisfying specific conditions, \eg, \myquote{How many objects share the 
same color as this one?},
(b) \textbf{Existence} questions are yes/no binary questions that verify
conditions in the image, \eg, \myquote{Are there any other cubes?}, and
(c) \textbf{Seek} questions query attributes of objects,
\eg, \myquote{What color is that cylinder?}. 

Consider \reffig{fig:question_generation} that shows how the current question is 
generated using the primitives and grammar, given the caption and dialog history 
(question-answer pair for the first three rounds).
For the current round, the question 
\textit{`What material is the green object at the back?'}
is clearly implausible (Q-er is unaware of the existence of a green object),
while the question \textit{`What shape is the red object?'} is redundant.
For the templates visualized, \texttt{Unique[object]} returns a list of unique
known object-attribute pairs (using $S^t_q$).
A candidate is sampled by \texttt{Sample[object]} and a relation is applied through
\texttt{Relate(in front of)}.
There are multiple choices at this junction:
(a) The use of \texttt{Count} leads to a counting question (\texttt{count-obj-rel-early}),
(b) Invoking \texttt{Sample[attribute]} results in a seek question (\texttt{seek-attr-rel-early}), and finally,
(c) \texttt{Exist} primitive generates an exist question of type \texttt{exist-obj-rel-early}.

\begin{table*}[!htbp]
	\centering
    \begin{tabular}[t]{r p{0.7\columnwidth}}
	\toprule
    %\textbf{Name} & \textbf{Example} \\
	%\midrule
    \multicolumn{2}{c}{\textbf{Captions}}\\
    \midrule
    %\cdashlinelr{1-3}
    \multirow{2}{*}{\texttt{obj-relation}}
        & \textit{`A [Z] [C] [M] [S] stands [R] a [Z1] [C1] [M1] [S1].'} \\
        & \textit{`A gray sphere stands to the right of a red object.'} \\
        
    \multirow{2}{*}{\texttt{obj-unique}}
        & \textit{'A [Z] [C] [M] [S] is present in the image.'}\\
        & \textit{`A red object is present in the image'}\\
        
    \multirow{2}{*}{\texttt{obj-extreme}}
        & \textit{`The rightmost thing in the view is a [Z] [C] [M] [S].'}\\
        & \textit{`The rightmost thing in the view is a cylinder.'}\\
        
    \multirow{2}{*}{\texttt{obj-count}}
        & \textit{`The image has [X] [Z] [C] [M] [S].'}\\
        & \textit{`The image has four cylinders.'}\\
    \midrule
    \multicolumn{2}{c}{\textbf{Count/Exist Question Type}}\\
    \midrule
    \texttt{count-all}
        & \textit{`How many objects in the image?'} \\
        
	\texttt{count/}
	    & \textit{`[How many $|$ Are there] other [Z] [C] [M] [S] in the picture?'} \\
	\texttt{exist-excl}
        & \textit{`[How many $|$ Are there] other cubes in the picture?'} \\ 
        
	\texttt{count/} &
		\textit{`[If present, how many $|$ Are there] [Z] [C] [M] [S] objects?'} \\
    \texttt{exist-attr} &
		\textit{`[If present, how many $|$ Are there] metallic objects?'} \\
		
    \texttt{count/} &
		\textit{`[How many $|$ Are there] [Z] [C] [M] [S] among them?'} \\
	\texttt{exist-attr-group} &
		\textit{`[How many $|$ Are there] blue cylinders among them?'} \\
		
	\texttt{count/} &
		\textit{`[How many $|$ Are there] things to its [R]?'} \\
    \texttt{exist-obj-rel-imm} &
		\textit{`[How many $|$ Are there] things to its right?'} \\
		
	\texttt{count/} &
		\textit{`How about to its [R]?'} \\
	\texttt{exist-obj-rel-imm2} &
		\textit{`How about to its left?'} \\	
		
	\texttt{count/} &
 		\textit{`[How many $|$ Are there] things [R] that [Z] [C] [M] [S]?'} \\
 	\texttt{exist-obj-rel-early} &
 		\textit{`[How many $|$ Are there] things in front of that shiny object?'} \\
 		
	\texttt{count/} &
 		\textit{`[How many $|$ Are there] things that share its [A]?'} \\
    \texttt{exist-obj-excl-imm} &
 		\textit{`[How many $|$ Are there] things that share its color?'} \\
 		
    \texttt{count/}
    	& \textit{`[How many $|$ Are there] things that are the same [A] as that [Z] [C] [M] [S]?'} \\
    \texttt{exist-obj-excl-early}
    	& \textit{`[How many $|$ Are there] things that are the same size as that round object?'} \\
	\midrule
    \multicolumn{2}{c}{\textbf{Seek Question Type}}\\  
    \midrule
    \multirow{2}{*}{\texttt{seek-attr-imm}}
        & \textit{`What is its [A]?'} \\
        & \textit{`What is its shape?'} \\

    \multirow{2}{*}{\texttt{seek-attr-imm2}}
		& \textit{`How about [A]?'} \\
		& \textit{`How about color?'} \\
		
	\multirow{2}{*}{\texttt{seek-attr-early}}
	    & \textit{`What is the [A] of that [Z] [C] [M] [S]?'} \\
		& \textit{`What is the shape of that shiny thing?'} \\
	
	\multirow{2}{*}{\texttt{seek-attr-sim-early}}
        & \textit{`What about the earlier [Z] [C] [M] [S]?'} \\
        & \textit{`What about the earlier box?'} \\
 		
	\multirow{2}{*}{\texttt{seek-attr-rel-imm}}
 		& \textit{`If there is a thing to its [R], what [A] is it?'} \\
 		& \textit{`If there is a thing to its right, what color is it?'} \\
 		
    \multirow{2}{*}{\texttt{seek-attr-rel-early}}
    	& \textit{`If there is a thing [R] that [Z] [C] [M] [S], what [A] is it made of?'}\\
    	& \textit{`If there is a thing in front of that shiny object, what material is it made of?'}\\
	\bottomrule
\end{tabular}
	\caption{Example templates for all the caption and question types used 
    to generate \ours dataset.
    For each type, we show both: 
    (a) a sample template with placeholders 
    (Z=size, C=color, M=material, S=shape, A=attribute, X=count, R=relation), and
    (b) a realization with placeholders filled with random values.}
	\label{tab:templates}
\end{table*}
%A list of templates with example entries used to generate our dialog dataset are given in \reftab{tab:templates}.

\begin{figure*}[t]
	\centering
    \begin{subfigure}[b]{0.65\textwidth}
        \caption{Distribution of caption (left) and question (right) categories.}
        \label{fig:cap_ques_category_distr}
        \includegraphics[width=\textwidth]{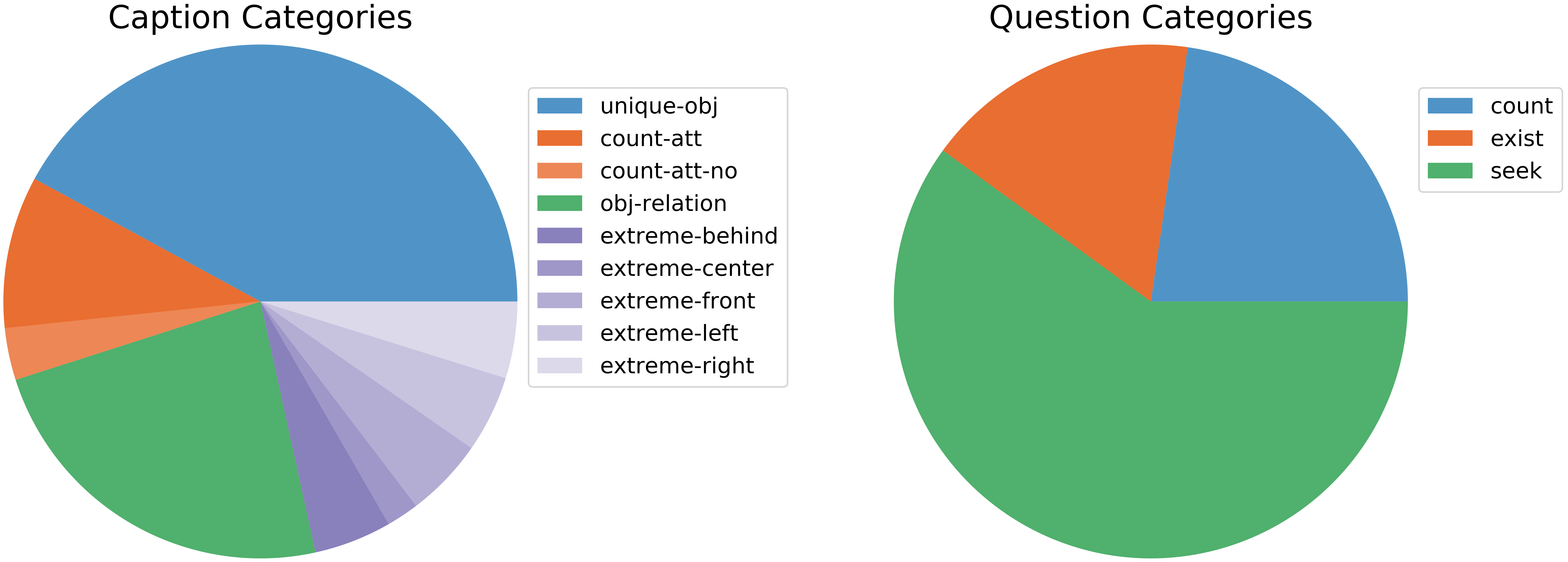}
    \end{subfigure}~
	\begin{subfigure}[b]{0.34\textwidth}
        \caption{Distribution of coreference distances.}
        \label{fig:coref_distr}
        \includegraphics[width=\textwidth]{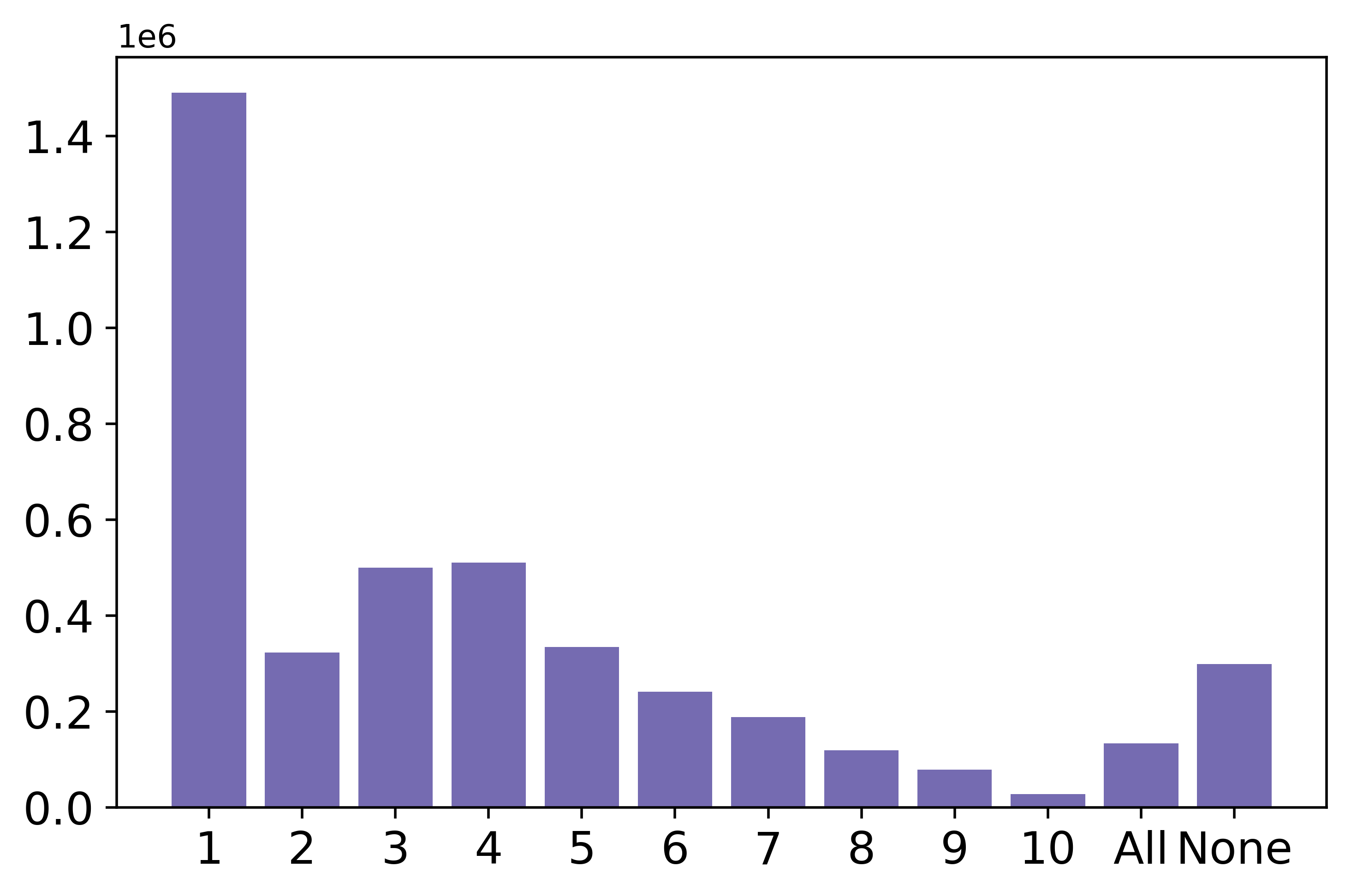}
    \end{subfigure}
    
    \begin{subfigure}[b]{0.99\textwidth}
        \caption{Distribution of questions according to the template labels.}
        \label{fig:ques_label_distr}
        \includegraphics[width=\textwidth]{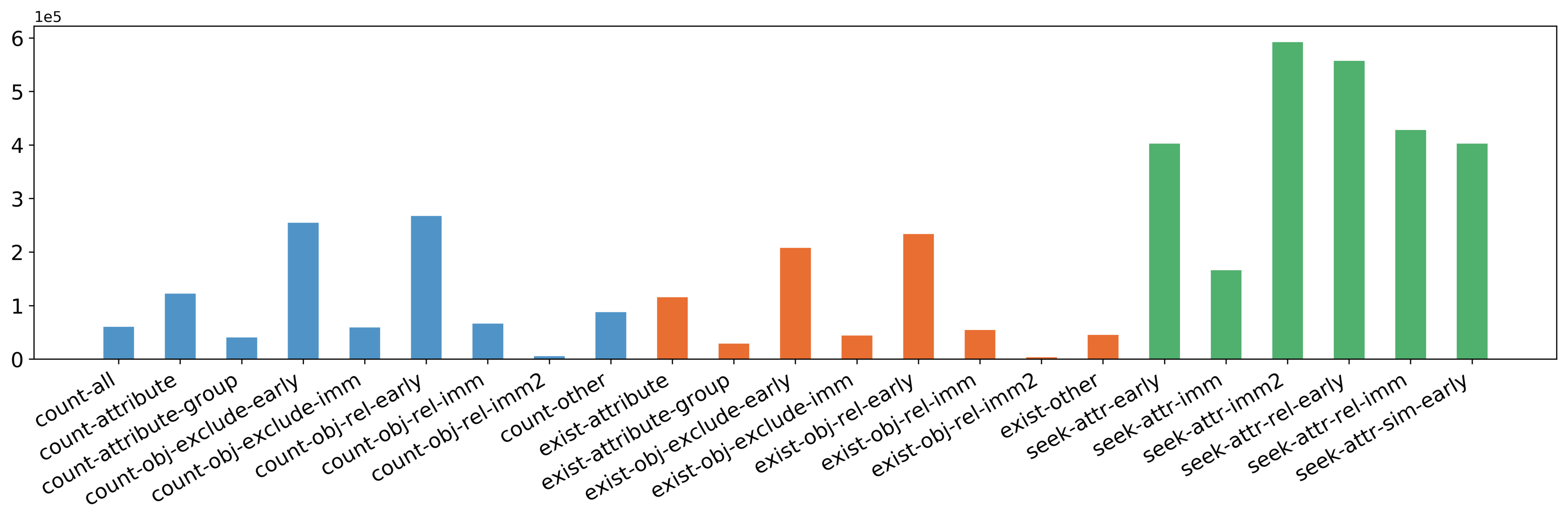}
    \end{subfigure}
    
    \begin{subfigure}[b]{0.99\textwidth}
        \caption{Distribution of answers.}
        \label{fig:ans_distr}
        \includegraphics[width=\textwidth]{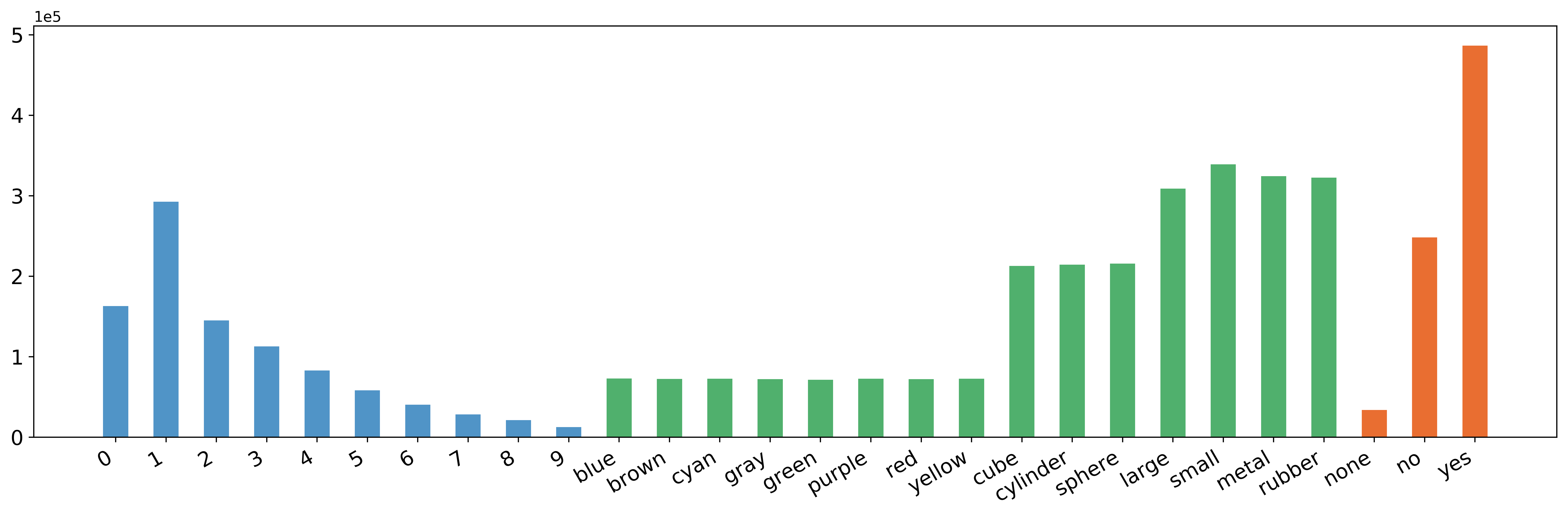}
    \end{subfigure}
    \caption{Visualization of various distributions for captions, questions, answers, and
    history dependency in our \ours dataset.
    See \refsec{sec:dataset_statistics} for more details.}
    \label{fig:additional_stats}
\end{figure*}

\paragraph{Dialog Generation.}
At a high level, dialog generation now `simply' involves selecting a 
sequence of templates such that the accompanying constraints are
satisfied by $S^t_q$ at all $t$. 
As a tractable approximation to this exponentially-large constraint satisfaction problem,
we use beam search that finds a valid solution \emph{and} enforces additional
conditions to make the dialog \textit{interesting}.
We found this to be effective both in terms of speed and dialog diversity.
More concretely, at every round of the dialog (after 3 rounds), we ensure that each of the 
question template types---count, existence, and seek---falls within a 
range ($10\%-30\%$ for count/existence each, and $30\%-60\%$ for seek)
%\marcus{[Marcus: Not clear how this can be a range? it has to add up to one hundred so it has to be 20+20+60 which will be difficult to satisfy?]}.
In addition, we identify \textit{independent} questions that do not need
history to answer them, \eg, \textit{`How many objects are present in the 
image?'}, and limit their number to under $20\%$.
Finally, to encourage questions that require reasoning over the history, \eg,
\texttt{seek-attr-sim-early} and \texttt{count-obj-excl-imm}, we tailor our
beam search objective so that dialogs containing such questions have a higher
value.
We use a beam search with $100$ beams for each dialog. 
\reffig{fig:dialog_generation} illustrates the diverse set of candidate questions
generated at each round for a given image.

To summarize, the usage of primitives and a dialog grammar makes our generation
procedure:
(a) modular: each primitive has an intuitive meaning,
(b) expressive: complex templates can be broken down into these primitives,
(c) computationally efficient: outputs can reused for templates sharing 
similar primitive structures (as seen in \reffig{fig:question_generation}), 
thus allowing an easy extension to new primitives and templates.
We believe that \ours represents not a static dataset but a recipe for 
constructing increasingly challenging grounded dialog by expanding this grammar.

%%%%%%%%%%%%%%%%%%%%%%%%%%%%%%%%%%%%%%%%%%%%%%%%%%%%%%%%%%%%%%%%%%%%%%%%%%%%%%%
\subsection{Dataset Statistics}
\label{sec:dataset_statistics}
We compare \ours to MNIST-Dialog and VisDial in \reftab{tab:statistics},
%table also contains VisDial for reference 
but the key 
measure of coreference distance cannot be reported for VisDial as it is not annotated. % in VisDial. 
Overall, \ours has $3 \times$ the questions and
a striking $206 \times$ the unique number of questions than MNIST-Dialog,
indicating higher linguistic diversity.
\ours questions are longer with a mean length of $10.6$ compared to
$8.9$ for MNIST-Dialog.
Crucially, supporting our motivation, 
the mean distance (in terms of rounds) between 
the coreferring expressions in \ours is $3.2\times$ compared to $1.0$ 
in MNIST-Dialog. Moreover, the distances in \ours vary 
(min of $1$, max of $10$), while it is constant (at 1) in 
MNIST-Dialog, making it easy for models to pick up on this bias. 

Further, we visualize the distribution of caption templates,
question templates, answers, and the history dependency of questions in \ours (\reffig{fig:additional_stats}), and discuss in detail below.

% statistics table
\begin{table}[t]
	\centering % \small
    \setlength{\tabcolsep}{2pt}
    \begin{tabular}[t]{l@{}ccc}
	\toprule
    \multirow{2}{*}{\textbf{Name}} & \textbf{CLEVR} & \textbf{MNIST} 
    	& \multirow{2}{*}{\textbf{VisDial}}\\
	& \textbf{Dialog (ours)} & \textbf{Dialog}\\
	\midrule
    $\#$ Images & $85k$ & $50k$ & $123k$\\
    $\#$ Dialogs & $425k$ & $150k$ & $123k$\\
    $\#$ Questions & $4.25M$ & $1.5M$ & $1.2M$\\
    $\#$ Unique Q & $73k$ & $355$ & $380k$\\
    $\#$ Unique A & $29$ & $38$ & $340k$\\
    Vocab. Size & 125 & 54 & $7k$\\
    Mean Q Len. & 10.6 & 8.9 & 5.1\\
    Mean Coref Dist. & 3.2 & 1.0 & -\\
    \bottomrule
\end{tabular}
	\caption{
    Dataset statistics comparing \ours to MNIST Dialog \cite{paul2017visual}.
    Our dataset has $3\times$ the questions (larger), 
    $206\times$ the unique number of questions (more diverse),
    $3.2\times$ the mean coreference distance (more complex),
    and longer question lengths. Similar stats for VisDial 
    shown for completeness.
    Coreference distance can not be computed for VisDial due to lack of annotations.\vspace*{-10pt}}
	\label{tab:statistics}
\end{table}

% \marcus{[Marcus: the following paragraph seems to belong not in this section on statistics but seems to fit more as extension to the "Caption Generation" paragraph in section 3.2!? (the last sentence might have to be omitted for that)}

\paragraph{Question Categories and Types.}
\ours contains three broad question categories---count, exist, and seek---with 
each further containing variants totaling up to $23$ different types of questions.
In comparison, MNIST-Dialog only has $5$ types of questions and is less diverse.
The distributions for the question categories and question types are shown in 
\reffig{fig:cap_ques_category_distr} and \reffig{fig:ques_label_distr}, respectively.
Our questions are $60\%$ seek as they open up more interesting follow-up questions,
$23\%$ count, and $17\%$ exist.

\paragraph{History Dependency.}
Recall that our motivation for \ours to create a diagnostic dataset for
multi-round reasoning in visual dialog.
As a result, a majority of questions in our dataset depend on the dialog history.
We identify three major kinds of history dependency for the questions:
(a) \textbf{Coreference} occurs when a phrase within the current question refers to a
earlier mentioned object (referent).
We characterize coreferences by measuring the distance between the current and
the earlier mention, in terms of dialog rounds.
This can range from $1$ (\eg., \textit{`What is its color?'}) to $10$
(a question in round $10$ referring to an entity in the caption).
(b) \textbf{All}: When the question depends on the entire dialog history,
\eg., \textit{`How many other objects are present in the image?'}, 
(c) \textbf{None}: When the question is stand-alone and does not depend on the history,
\eg., \textit{`How many spheres does the scene have?'}
The distribution of questions characterized according to the history dependency is
shown in \reffig{fig:coref_distr}.
Unlike MNIST Dialog, \ours contains a good distribution of reference distances beyond
just $1$, leading to a mean distance of $3.2$.
Thus, the models will need to reason through different rounds of dialog history 
in order to succeed.
%%%%%%%%%%%%%%%%%%%%%%%%%%%%%%%%%%%%%%%%%%%%%%%%%%%%%%%%%%%%%%%%%%%%%%%%%%%

\section{Experiments}
In this section, we describe and benchmark several models on \ours.
We then breakdown and analyze their performance according to question type
and history dependency.
Finally, we focus on the best performing model and study its behavior 
on \ours both qualitatively and quantitatively.
Specifically, we visualize qualitative examples and develop metrics to 
quantitatively evaluate the textual and visual grounding.
Note that such a diagnostic analysis of visual dialog models is first of 
its kind which would not be possible without our \ours.

\subsection{Baselines}
To benchmark performance, we evaluate several models on \ours. 
\textbf{Random} picks an answer at random.
\textbf{Random-Q} picks an answer at random among valid answers for a given question type (\eg, name of a color for color questions).
Further, we adapt the discriminative visual dialog models from
\citet{visdial}:
(a) \textbf{L}ate \textbf{F}usion (\textbf{LF}) that models 
separately encode each of question (Q), history (H), and image (I); and then
fuse them by concatenation.
(b) \textbf{H}ierarchical \textbf{R}ecurrent \textbf{E}ncoder (\textbf{HRE}) that models 	
dialog via both dialog-level and sentence-level recurrent neural networks.
(c) \textbf{M}emory \textbf{N}etwork (\textbf{MN}) that stores history as memory
units and retrieves them based on the current question.
We also consider neural modular architectures:
(a) \textbf{CorefNMN} \cite{Kottur_2018_ECCV} that explicitly models coreferences in 
visual dialog by identifying the \textit{reference} in the question (textual grounding)
and then localizing the \textit{referent} in the image (visual grounding), and
(b) \textbf{NMN} \cite{hu2017learning}, which is a history-agnostic ablation of CorefNMN.

\begin{SCtable}
	\centering
    \begin{tabular}[t]{lc}
	\toprule
	\textbf{Model} & \textbf{Acc.}\\
	\midrule
    Random & 3.4 \\
    Random-Q & 33.4 \\
    \cdashlinelr{1-2}
    LF-Q & 40.3 \\
    LF-QI & 50.4 \\
    LF-QH & 44.1 \\
    LF-QIH & 55.9 \\
    \cdashlinelr{1-2}
    HRE-QH & 45.9 \\
    HRE-QIH & 63.3 \\
    \cdashlinelr{1-2}
    MN-QH & 44.2 \\
    MN-QIH & 59.6 \\
    \cdashlinelr{1-2}
    NMN & 56.6 \\
    %NMN(O) & xx.x \\
    %\cdashlinelr{1-2}
    CorefNMN & \textbf{68.0} \\
    %CorefNMN(O) & xx.x\\
    \bottomrule
\end{tabular}
	\caption{
    Accuracy ($\%$) on \ours (higher is better).
    See text for details.\vspace*{-5pt}}
	\label{tab:results}
\end{SCtable}

% Split this into results table + breakdown figure.

\subsection{Overall Results}
	We use multi-class classification accuracy for evaluation 
    since \ours has one-word answers. 
	\reftab{tab:results} shows the performance of different models. % on \ours.
    The key observations are:
    (a) Neural models outperform random baselines by a large margin.
    The best performing model, CorefNMN, outperforms Random-Q by 35\%.
    (b) As expected, blind models (LF-Q, LF-QH, HRE-QH, MN-QH)
    are inferior to their counterparts that use I, by at least 10\%.
    (c) History-agnostic models (LF-Q, LF-QI, NMN) also suffer
    in performance, highlighting the importance of history.

%%%%%%%%%%%%%%%%%%%%%%%%%%%%%%%%%%%%%%%%%%%%%%%%%%%%%%%%%%%%%%%%%%%%%%%%%%%
\subsection{Accuracy vs History Dependency}
\begin{figure}[t]
    \centering
    \includegraphics[width=0.98\columnwidth]{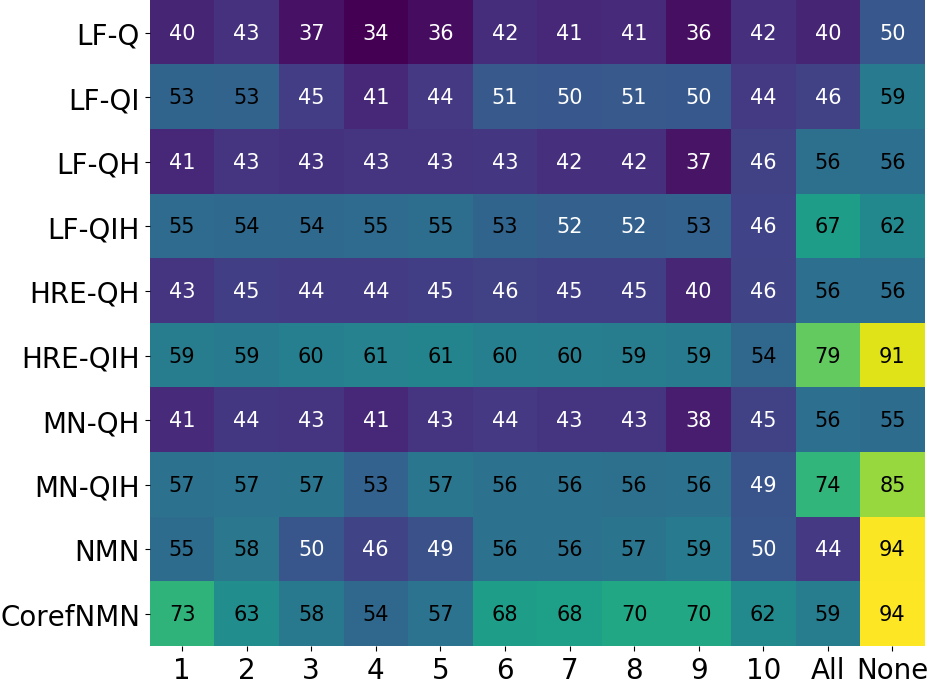}
    \caption{Accuracy breakdown of models according to the history dependency type.
    While CorefNMN outperforms all methods on questions (average) containing references
    ($1-10$), 
    its performance is not as good on questions that depend on the 
    entire history (`All').}
    \label{fig:accuracy_coref_dependency}
\end{figure}

The breakdown of model performances based on the history dependency is presented in
\reffig{fig:accuracy_coref_dependency}.
The following are the important observations:
\begin{itemize}[leftmargin=0.15in,itemsep=-5pt]
    \item The best performing model, CorefNMN, has a superior performance
    (on an average) on all question with coreference ($1-10$) compared to
    all other models.
    As CorefNMN is designed specifically to handle coreferences in visual dialog,
    this is not surprising.
    \item Interestingly, the second best model HRE-QIH has the best accuracy
    on `All' questions, even beating CorefNMN by a margin of $20\%$.
    In other words, HRE-QIH (and even MN-QIH) is able to answer `All'
    questions significantly better than CorefNMN perhaps due to the ability
    of its dialog-level RNN to summarize information as the dialog progresses.
    \item Both NMN and CorefNMN perform similarly on the `None' questions.
    This observation is intuitive as NMN is a history-agnostic version of 
    CorefNMN by construction.
    However, the difference becomes evident as CorefNMN outperforms NMN by
    about $12\%$ overall.
\end{itemize}
%%%%%%%%%%%%%%%%%%%%%%%%%%%%%%%%%%%%%%%%%%%%%%%%%%%%%%%%%%%%%%%%%%%%%%%%%%%

%%%%%%%%%%%%%%%%%%%%%%%%%%%%%%%%%%%%%%%%%%%%%%%%%%%%%%%%%%%%%%%%%%%%%%%%%%%
\begin{figure*}[t]
    \centering
    \includegraphics[width=\columnwidth]{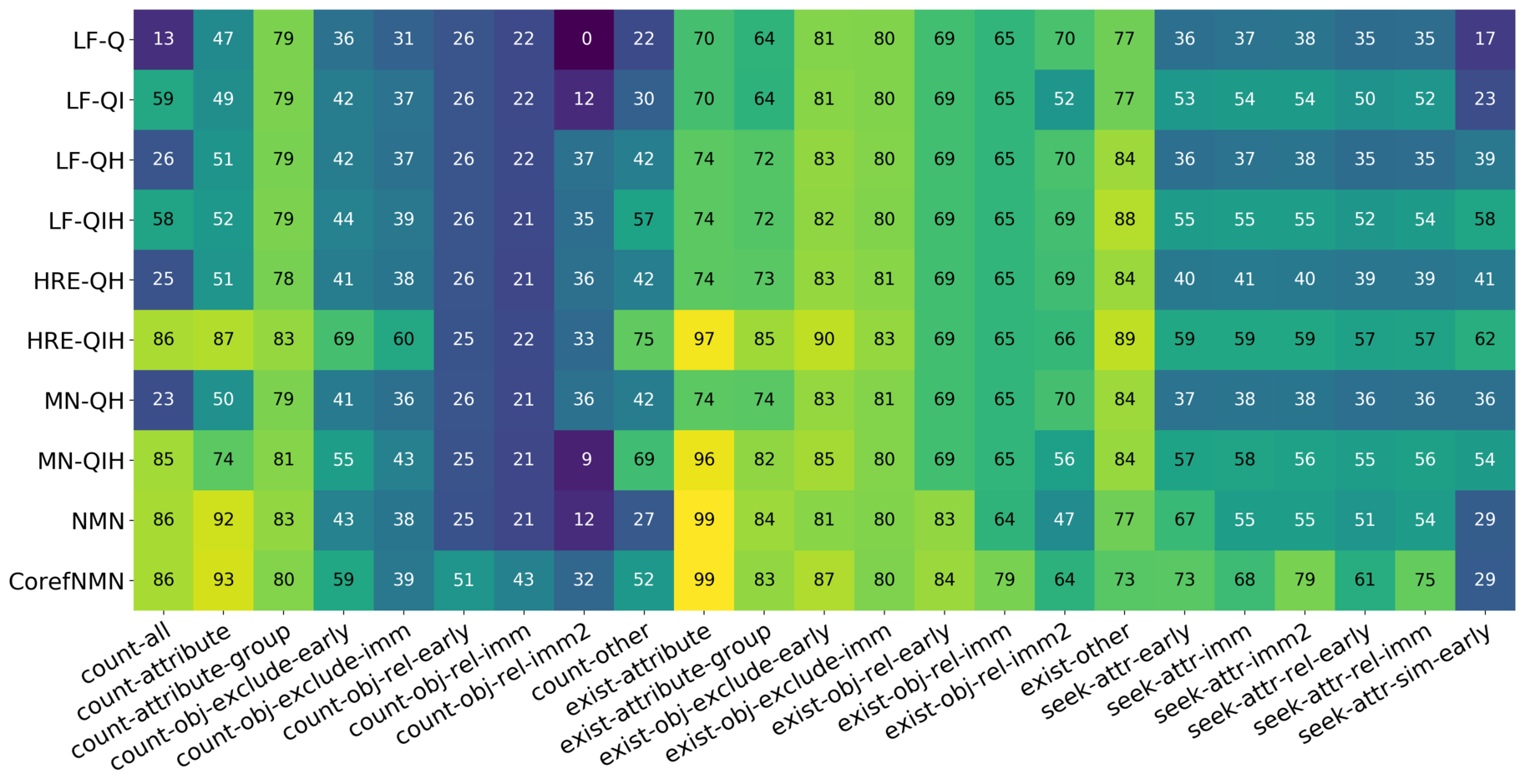}
    \caption{Accuracy breakdown of models according to the question type.
            See text in \refsec{sec:accuracy_question_type} for more details.}
    \label{fig:accuracy_qtype}
\end{figure*}
%%%%%%%%%%%%%%%%%%%%%%%%%%%%%%%%%%%%%%%%%%%%%%%%%%%%%%%%%%%%%%%%%%%%%%%%%%%
\begin{figure}[t]
    \centering
    \includegraphics[width=0.99\columnwidth]{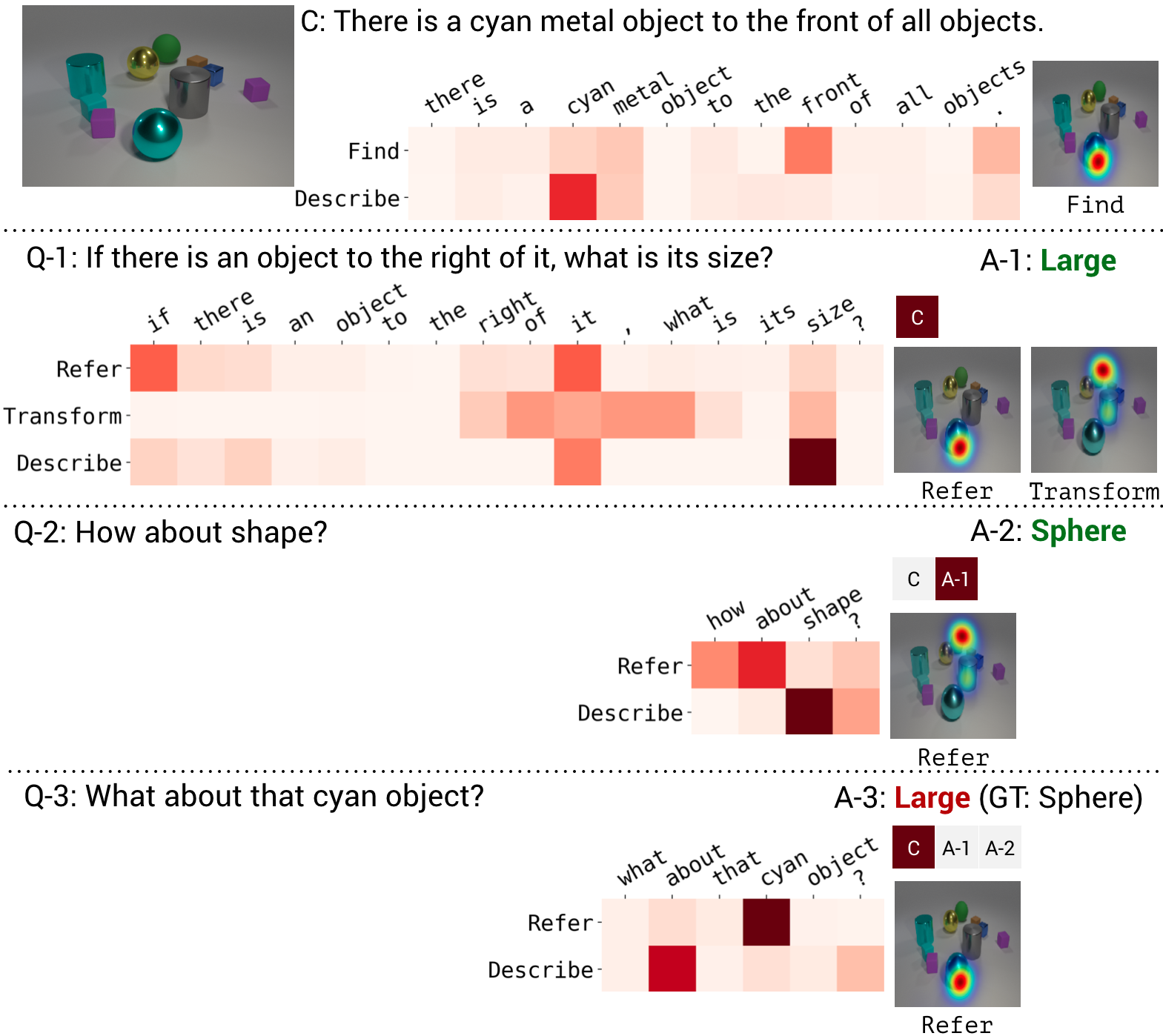}
    \caption{Qualitative visualization of CorefNMN on \ours.}
    \label{fig:corefnmn_qual}
\end{figure}

\subsection{Accuracy vs Question Type}
\label{sec:accuracy_question_type}
\reffig{fig:accuracy_qtype} breaks down the performance of all the models according
to the question types.
An obvious observation is that performance on counting and seek questions is worse 
than that on exist questions.
While this is in part because of the binary nature of exist questions, they are
also easier to answer than counting or extracting attributes that need complicated 
visual understanding.
%%%%%%%%%%%%%%%%%%%%%%%%%%%%%%%%%%%%%%%%%%%%%%%%%%%%%%%%%%%%%%%%%%%%%%%%%%%

\subsection{Qualitative Anaylsis for CorefNMN}
We now qualitatively visualize (\reffig{fig:corefnmn_qual}) the 
best performing model, CorefNMN.
%, in a format similar to \newcite{Kottur_2018_ECCV}.\marcus{[I would remove the reference, this seems to more clearly point to that we are the same authors.]}
In the example shown, CorefNMN first parses the caption 
\textit{`There is a cyan metal object to the front of all the objects.'} and 
localizes the right cyan object.
While answering Q-1, CorefNMN rightly instantiates the \texttt{Refer} module
and applies the desired transformation (see module outputs on the right).
For Q-2, it accurately identifies the object as the previous one,
and extracts the attributes.
Finally, the question \textit{`What about that cyan object?'} cannot be
answered in isolation as: (a) there are multiple cyan objects,
(b) the meaning of the question is incomplete without Q-2.
It is interesting to note that even though CorefNMN overcomes (a)
by correctly resolving the reference 
\textit{that cyan object} (in the image), it is unable to 
circumvent (b) due to its specialization in visual coreferences.

We also provide additional analysis to evaluate the textual and
visual grounding by CorefNMN in the supplement.

% \input{sections/app/corefnmn_qual_examples.tex}
%%%%%%%%%%%%%%%%%%%%%%%%%%%%%%%%%%%%%%%%%%%%%%%%%%%%%%%%%%%%%%%%%%%%%%%%%%%

%\section{Model Analysis}
% \vspace{-10pt}
%\input{sections/main/model_analysis.tex}

\section{Conclusion}
We proposed a large, synthetic dataset called \ours,
to study multi-round reasoning in visual dialog, and in particular
the challenge of visual coreference resolution.
We benchmarked several qualitatively different models from prior work 
on this dataset, which act as baselines for future work.
Our dataset opens the door to evaluate how well models do on 
visual coreference resolution, 
without the need to collect expensive annotations on real 
datasets.
% \marcus{\section{Final todos}}
% \begin{itemize}
%     \item make a pass over references and check for duplicates and see if arxives are publications, e.g. "Improving coreference resolution by learning entity-level distributed representations." was at ACL 16 I think?
%     % \item remove comments and color before submitting to arxiv
% \end{itemize}

\clearpage
\section*{Supplementary}
\appendix
The supplement is organized as follows:
\begin{itemize}
    \item Grounding analysis for the best performing model, CorefNMN, in 
    \refsec{app:grounding_analysis_corefnmn},
    \item \refsec{app:implementation_details} provides implementation details.
\end{itemize}

\section{Grounding Analysis for CorefNMN}
\label{app:grounding_analysis_corefnmn}
As mentioned earlier, CorefNMN identifies a reference phrase in
the current question and proceeds to visually ground the corresponding referent
in the image.
Such explicit textual and visual grounding at each round allows for
an interesting quantitative analysis for CorefNMN, with the help of
annotations in our \ours.
% Essentially, \ours provides coreference annotations for each question
% in the form of a reference phrase, if any, and its bounding box localization in the image.
% By comparing these grounding annotations with the output from the model, we can
% quantitatively assess grounding (both textual and visual) by CorefNMN.
In what follows, we first describe the grounding annotations, detail the evaluation 
procedure, and then present our observations.

\begin{figure*}[t]
	\centering
    
    \begin{subfigure}[b]{0.99\textwidth}
        \caption{NDCG value for text grounding for various question types.}
        \label{fig:corefnmn_ndcg_text}
        \includegraphics[width=\textwidth]{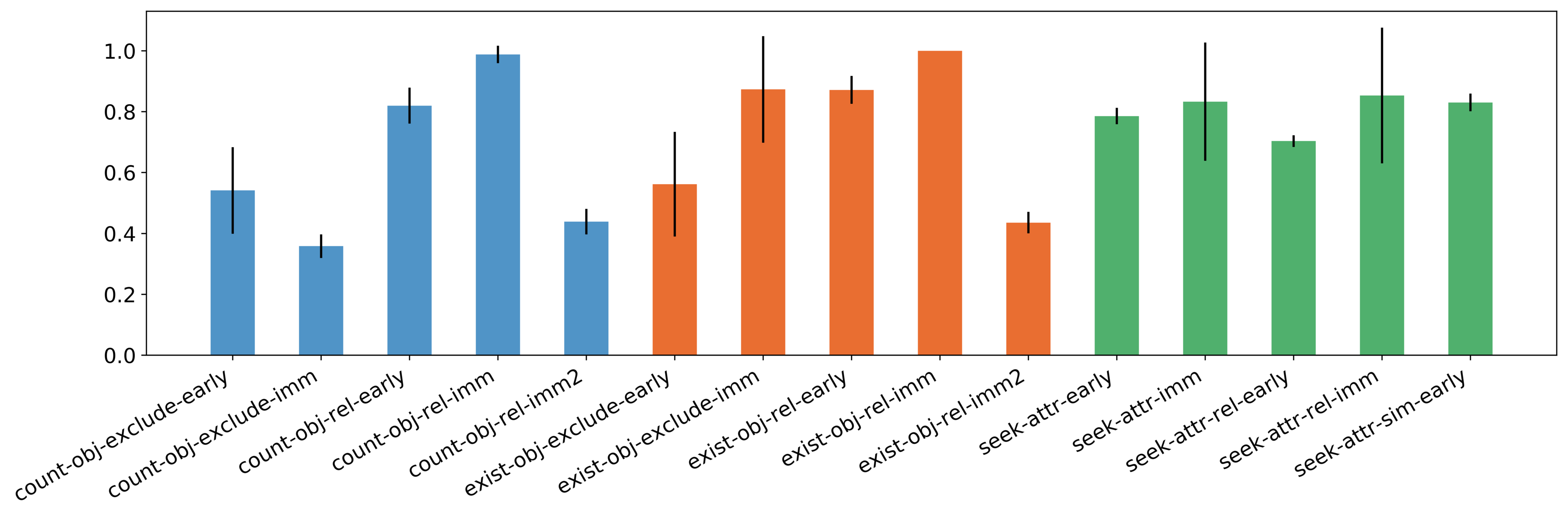}
    \end{subfigure}
    
    \begin{subfigure}[b]{0.99\textwidth}
        \caption{NDCG value for visual grounding for various question types.}
        \label{fig:corefnmn_ndcg_image}
        \includegraphics[width=\textwidth]{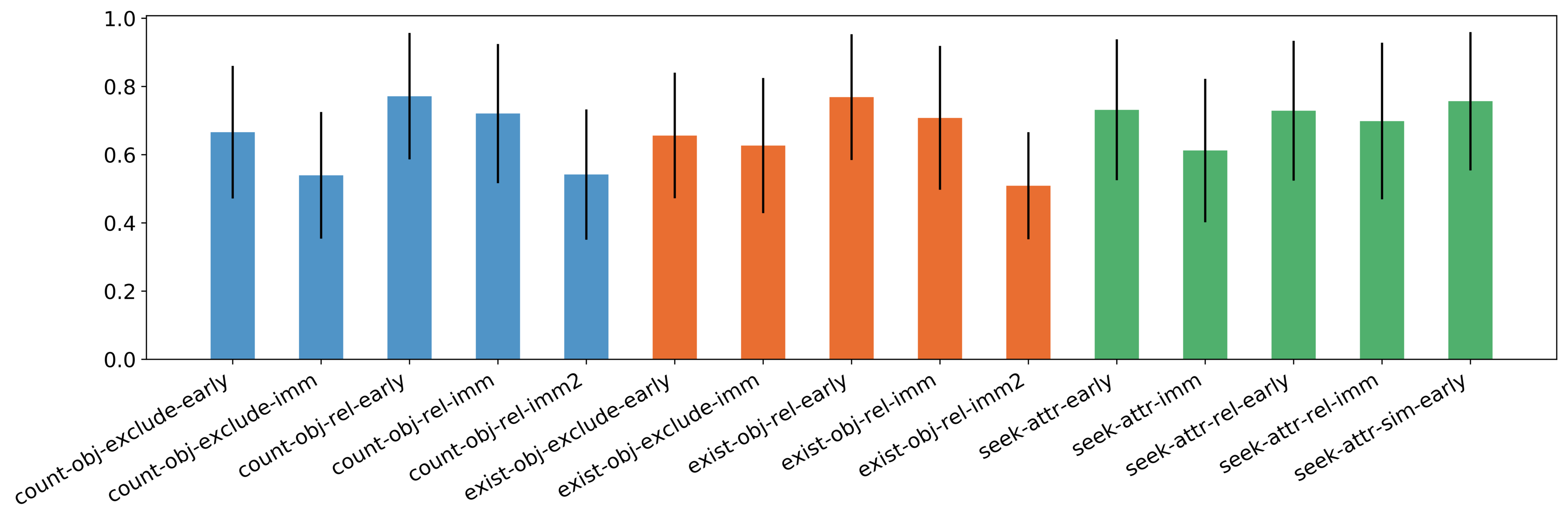}
    \end{subfigure}
    
    \caption{Evaluating the textual (above) and visual (below) grounding of CorefNMN on \ours, using Normalized Discounted Cumulative Gain (NDCG) for various question
    types. Higher is better.}
    \label{fig:corefnmn_grounding_metrics}
\end{figure*}

\paragraph{Annotations.}
While the original CLEVR dataset \cite{johnson2017clevr} does not contain bounding 
box annotations for the objects in the scene,
\citet{krishna2018referring} later added these in their
work on referring expressions.
We leverage these annotations to obtain the ground truth visual groundings ($A_V$) for 
the referents in our questions.
On the other hand, each of the caption and question templates has referring phrase 
annotations in them, thus giving the ground truth textual groundings ($A_T$).
We use the above two groundings for evaluation.

\paragraph{Evaluation.}
For every coreference resolution, CorefNMN produces a visual attention map of size 
$14 \times 14$ ($\hat{{A}_V}$) and a textual attention over the question words ($\hat{{A}_T}$).
We rank all the $14^2=196$ cells in $\hat{{A}_V}$ according to their attention values.
Next, we appropriately scaled down $A_V$ ($14 \times 14$) and consider the cells
spanning the bounding box as relevant.
To evaluate grounding, we measure the retrieval performance of the relevant cells
in the sorted $\hat{A_V}$ through the widely used the
Normalized Discounted Cumulative Gain (NDCG)\footnote{\url{https://en.wikipedia.org/wiki/Discounted_cumulative_gain}} metric.
It is a measure of how highly the relevant cells were ranked in the sorted $\hat{{A}_V}$,
with a logarithmic weighting scheme to higher ranks, thus higher is better.
For the textual grounding, we perform a similar computation between $\hat{{A}_T}$
and $A_T$ and report NDCG.

\paragraph{Observations.}
The NDCG values to evaluate both textual and visual groundings for CorefNMN
are shown in \reffig{fig:corefnmn_grounding_metrics}.
An important takeaway is that the model is able to accurately ground the
references in the question (\reffig{fig:corefnmn_ndcg_text})
consistently for several question types, as reflected in a higher 
average NDCG.
Similarly, the visual grounding in \reffig{fig:corefnmn_ndcg_image} 
(average NDCG of $0.7$)
is significantly superior to a random baseline (NDCG of ~$0.3$).
%\marcus{what is the average performance as a number, not just bar plots?}
%\marcus{should we add a table for this?}
%\marcus{What about also evaluating grounding (not just coref?)}

\section{Implementation Details}
\label{app:implementation_details}
The dataset generation was done entirely in Python, without any significant
package dependencies.
To evaluate the models from \newcite{visdial}, we use their open source
implementation\footnote{\url{https://github.com/batra-mlp-lab/visdial}}
based on Lua Torch\footnote{\url{http://torch.ch/}}.
For the neural module architectures \cite{hu2017learning,Kottur_2018_ECCV}, we use the authors' Python-based,
publicly available 
implementations---NMN\footnote{\url{https://github.com/ronghanghu/n2nmn}} and
CorefNMN\footnote{\url{https://github.com/facebookresearch/corefnmn}}.
Questions are encoded by first learning a $128$-dimensional embedding for 
the words, which are then fed into a single layer LSTM of hidden size $128$.
We use a pretrained convolution neural network, ResNet-101 \cite{he16cvpr},
to extract features for the images.
Adam \cite{kingma2014adam} steps with a learning rate of $0.0001$ are
employed to maximize the log-likelihood of the ground truth answer, while
training.
A subset ($500$ images) of the training set is set aside to pick
the best performing model via early stopping.

\section{Document Changelog}
To help the readers track changes to this document, a brief changelog describing the revisions 
is provided below:

\textbf{v1:} NAACL 2019 submission.

\textbf{v2:} Added links to dataset and code.

\bibliographystyle{acl_natbib}
\bibliography{strings,biblioLong,rohrbach,main}

\end{document}